\documentclass{article}

\usepackage{arxiv}

\usepackage[utf8]{inputenc} % allow utf-8 input
\usepackage[T1]{fontenc}    % use 8-bit T1 fonts
\usepackage{hyperref}       % hyperlinks
\usepackage{url}            % simple URL typesetting
\usepackage{booktabs}       % professional-quality tables
\usepackage{amsfonts}       % blackboard math symbols
\usepackage{nicefrac}       % compact symbols for 1/2, etc.
\usepackage{microtype}      % microtypography
\usepackage{lipsum}
\usepackage{graphicx}
\graphicspath{ {./images/} }
\usepackage{amsmath}

\title{A Lightweight Multi-Module Fusion Approach for Korean Character Recognition}

\author{
  Inho Jake Park\thanks{First author, Formerly with GIST Laboratory Autonomous Driving (GLAD), Gwangju Institute of Science and Technology (GIST), South Korea} \\
  Computer Vision AI Research Team \\
  Among Solution\\ 
  Changwon, South Korea \\
  \texttt{inhopark2412@among.co.kr}
  \and
  \textbf{Jaehoon Jay Jeong}\thanks{Second Author} \\
  Computer Vision AI Research Team \\
  Among Solution \\
  Changwon, South Korea \\
  \texttt{jhjeong2409@among.co.kr}
  \and
  \textbf{Ho-sang Jo}\footnotemark[2] \\
  CEO \\
  Among Solution \\
  Changwon, South Korea \\
  \texttt{among@among.co.kr}
}

\begin{document}
\maketitle
\begin{abstract}
Optical Character Recognition (OCR) is crucial in various applications, such as document analysis, automated license plate recognition, and intelligent surveillance. However, traditional OCR models struggle with irregular text structures, low-quality inputs, character variations, and high computational costs, making them unsuitable for real-time and resource-constrained environments.

In this paper, we introduce \textbf{Stroke-Sensitive Attention and Dynamic Context Encoding Network (SDA-Net)}, a novel architecture designed to enhance OCR performance while maintaining computational efficiency. Our model integrates:
\begin{itemize}
    \item a \textbf{Dual Attention Mechanism (DAM)} consisting of Stroke-Sensitive Attention and Edge-Aware Spatial Attention to improve stroke-level representation,
    \item a \textbf{Dynamic Context Encoding (DCE)} module to refine contextual information through a learnable gating mechanism,
    \item an \textbf{Efficient Feature Fusion Strategy} inspired by U-Net, which enhances character representation by combining low-level stroke details with high-level semantic information,
    \item and an \textbf{Optimized Lightweight Architecture} that significantly reduces memory usage and computational overhead while preserving accuracy.
\end{itemize}

Experimental results demonstrate that SDA-Net outperforms existing methods on multiple challenging OCR benchmarks while achieving faster inference speeds, making it well-suited for real-time OCR applications on edge devices.
\end{abstract}

% keywords can be removed

\section{Introduction}

Optical Character Recognition (OCR) has been widely applied in various fields, including automated text extraction, license plate recognition, and real-time surveillance. Despite significant advancements in deep learning-based OCR, current models often face challenges in recognizing characters under real-world conditions, such as:

\begin{itemize}
    \item \textbf{Stroke-Level Distortions}: Many OCR systems fail to capture fine-grained stroke information, leading to misclassification in handwritten or degraded text.
    \item \textbf{Contextual Ambiguities}: Context information is often ignored or statically encoded, limiting the model’s ability to infer missing or occluded characters.
    \item \textbf{Weak Feature Fusion}: Most models do not effectively integrate low-level and high-level representations, resulting in suboptimal performance.
    \item \textbf{Computational Inefficiency}: Many existing OCR models rely on heavy computation, making them impractical for edge devices or real-time applications.
\end{itemize}

To address these issues, we propose the \textbf{Stroke-Sensitive Attention and Dynamic Context Encoding Network (SDA-Net)}, which introduces:

\begin{enumerate}
    \item \textbf{Stroke-Sensitive Attention}: A novel attention mechanism that enhances character stroke perception, improving recognition accuracy in noisy environments.
    \item \textbf{Dynamic Context Encoding}: A lightweight encoding module that dynamically refines feature representations using a learnable gating mechanism.
    \item \textbf{Feature Fusion with Skip Connections}: Inspired by U-Net, our model fuses low-level stroke information with high-level semantic features, ensuring comprehensive character representation.
    \item \textbf{Efficient Model Design}: We optimize the network architecture to reduce computational overhead while maintaining high accuracy, making it suitable for real-time and resource-constrained environments.
\end{enumerate}

\subsection{Key Contributions}

This paper presents the following key contributions:

\begin{itemize}
    \item We introduce a \textbf{Dual Attention Mechanism} that integrates stroke-level attention with spatial edge-aware attention, enhancing fine-grained text representation.
    \item We propose a \textbf{Dynamic Context Encoding} module that adaptively refines feature weights to improve OCR performance.
    \item We develop an efficient \textbf{Feature Fusion Strategy} that combines multi-scale representations, improving robustness in challenging conditions.
    \item We optimize the architecture to achieve a \textbf{lightweight design} with reduced memory consumption and computation, ensuring fast inference.
    \item We evaluate our model on multiple OCR benchmarks and demonstrate \textbf{state-of-the-art performance} in noisy, occluded, and low-resolution text recognition scenarios.
\end{itemize}

\section{Related Works}

Optical Character Recognition (OCR) has seen significant advancements through deep learning-based methods. Traditional OCR systems relied on handcrafted features and rule-based approaches~\cite{ref11}, which struggled in recognizing complex scripts, noisy backgrounds, and low-resolution text. With the emergence of deep learning, several attention-based architectures have improved text recognition performance.

\subsection{Attention-Based OCR Models}

Attention mechanisms have played a crucial role in improving OCR accuracy. ASTER~\cite{ref11} and SAR~\cite{ref12} introduced spatial attention to focus on relevant regions of the text, but they lacked fine-grained stroke sensitivity, leading to misclassification in degraded or handwritten text. Transformer-based approaches like SATRN~\cite{ref13} and TrOCR~\cite{ref17} improved global context modeling but required large-scale datasets and suffered from high computational costs.

Recent models, such as VisionLAN~\cite{ref14} and SEED~\cite{ref15}, introduced global-local attention mechanisms and semantic reasoning, respectively, to enhance contextual awareness. However, these models still rely on static context encoding, limiting their adaptability to occlusions and missing characters. MASTER~\cite{ref16} proposed multi-scale attention but lacked explicit feature fusion strategies for integrating stroke-level information.

\subsection{Lightweight OCR Models}

To optimize OCR for mobile and real-time applications, lightweight models like PP-OCRv3~\cite{ref19} have been developed. PP-OCRv3 employs a combination of efficient attention mechanisms and implicit feature fusion to reduce computational costs. However, it sacrifices fine-grained stroke sensitivity and contextual adaptability. Similarly, EasyOCR~\cite{ref20} relies on LSTM-based sequence encoding without explicit attention, making it less effective in complex text recognition tasks.

\subsection{Proposed SDA-Net}

To address these limitations, we propose the Stroke-Sensitive Attention and Dynamic Context Encoding Network (SDA-Net). Unlike existing models, SDA-Net introduces:

\begin{itemize}
    \item \textbf{Stroke-Sensitive Attention (SSA)}: Captures fine-grained stroke details, improving robustness in noisy and occluded environments.
    \item \textbf{Edge-Aware Attention}: Enhances spatial structure awareness for better text boundary perception.
    \item \textbf{Dynamic Context Encoding (DCE)}: Implements a learnable gating mechanism to adaptively refine feature representations.
    \item \textbf{Explicit Feature Fusion (U-Net Inspired)}: Ensures effective integration of low-level stroke details with high-level semantic information.
\end{itemize}

SDA-Net significantly improves OCR accuracy while maintaining a lightweight design (5.6M parameters, 3.4 GFLOPs), making it an optimal balance between efficiency and performance. Our method demonstrates superior recognition on challenging datasets compared to existing approaches, particularly in scenarios involving distorted, occluded, and low-resolution text.

\begin{table}[htbp]
\centering
\caption{Comparison of OCR models including EasyOCR and the proposed SDA-Net.}
\label{tab:ocr_comparison}
\resizebox{\textwidth}{!}{
\begin{tabular}{|l|c|c|c|c|c|}
\hline
\textbf{Model} & \textbf{Year} & \textbf{Attention Type} & \textbf{Context Encoding} & \textbf{Feature Fusion} & \textbf{Params (M)} \\
\hline
ASTER~\cite{ref11} & 2018 & Seq-to-Seq Attention (LSTM) & BiLSTM Encoder & None & 27.2 \\
SAR~\cite{ref12} & 2019 & 2D Spatial Attention & Self-Attention (No RNN) & None & 27.8 \\
SATRN~\cite{ref13} & 2020 & Transformer-based 2D Attention & Implicit (Transformer) & None & -- \\
VisionLAN~\cite{ref14} & 2021 & Integrated Visual-Language Attention & Implicit (Context within Visual Features) & Implicit Fusion & 42.2 \\
SEED~\cite{ref15} & 2020 & Sequence Attention + Semantic Guidance & BiLSTM + Semantic Prediction & None & 36.1 \\
MASTER~\cite{ref16} & 2021 & Multi-head Self-Attention (Transformer) & Implicit Global Context & Multi-Aspect Fusion & 62.8 \\
TrOCR~\cite{ref17} & 2021 & Transformer Encoder-Decoder & ViT Encoder + Text Decoder & None & 83.9 \\
DTrOCR~\cite{ref18} & 2023 & Decoder-only Transformer (GPT-like) & Implicit (Pretrained LM) & None & 105 \\
PP-OCRv3~\cite{ref19} & 2022 & None (CTC-based with SVTR module) & Implicit (SVTR-LCNet) & None & 12.4 \\
EasyOCR~\cite{ref20} & 2020 & None (CNN+LSTM+CTC) & BiLSTM Encoder & None & 8.7 \\
\textbf{Ours (SDA-Net)} & 2025 & Stroke-Aware + Edge Attention & Dynamic (Learnable Gating) & U-Net Style Fusion & \textbf{5.6} \\
\hline
\end{tabular}
}
\end{table}
\section{Method}

In this section, we present our proposed model for single-character OCR recognition. Our network is designed to capture both low-level details and high-level contextual information by integrating a ResNet-based feature extractor, a dual attention module, a dynamic context encoding module, and a fusion mechanism that combines these multi-scale features. The following subsections describe each component in detail.

\subsection{Overall Architecture}

Given an input image $I \in \mathbb{R}^{B \times 3 \times H \times W}$, our model extracts robust visual representations using a combination of ResNet-based Feature Extraction, Dual Attention Mechanism, Dynamic Context Encoding, and Feature Fusion with Skip Connection:

\begin{figure}[ht!]
\centering
\includegraphics[width=0.9\textwidth]{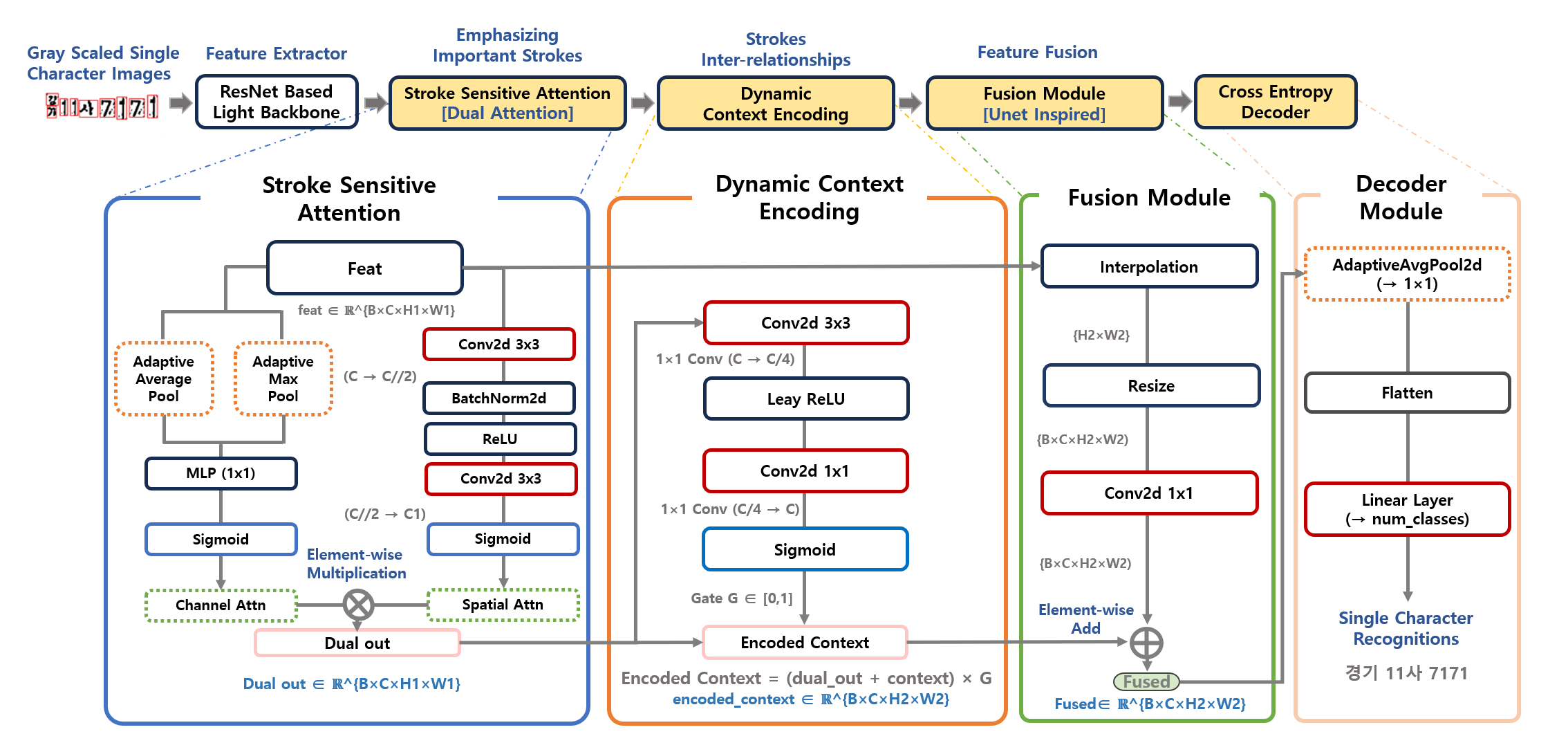}
\caption{Proposed SDA-Net architecture}
\label{fig: Proposed SDA-Net architecture visualization}
\end{figure}

\begin{equation}
F = E(I, \theta_E)
\end{equation}

where $E$ denotes the feature extractor parameterized by $\theta_E$, and $F \in \mathbb{R}^{B \times C \times H' \times W'}$ is the extracted feature map.

\subsection{Feature Extraction}

The ResNet-based feature extractor produces a feature representation $F$ by applying multiple residual blocks:

\begin{equation}
F = \text{ResNet}(I)
\end{equation}

Each layer performs:

\begin{equation}
F_{\ell+1} = \sigma(W_\ell * F_\ell + b_\ell)
\end{equation}

where $W_\ell$ is the convolution kernel, $*$ denotes convolution, and $\sigma$ is a ReLU activation.

\subsection{Dual Attention Mechanism}

To improve feature selectivity, we apply a Dual Attention Mechanism consisting of Channel Attention and Spatial (Edge) Attention.

\subsubsection{Channel Attention}

\begin{equation}
A_{\text{chan}} = \sigma(W_c(\text{MLP}(\text{AvgPool}(F)) + \text{MLP}(\text{MaxPool}(F))))
\end{equation}

\begin{equation}
F_{\text{chan}} = F \odot A_{\text{chan}}
\end{equation}

where MLP is defined as $\text{MLP}(x) = W_2(\text{ReLU}(W_1 x + b_1)) + b_2$, and $\odot$ denotes element-wise multiplication.

\subsubsection{Spatial Attention (Edge Attention)}

\begin{equation}
A_{\text{spat}} = \sigma(W_s * \text{ReLU}(W_e * F + b_e))
\end{equation}

\begin{equation}
F_{\text{spat}} = F \odot A_{\text{spat}}
\end{equation}

\begin{equation}
F_{\text{dual}} = F_{\text{chan}} + F_{\text{spat}}
\end{equation}

\subsection{Dynamic Context Encoding}

To capture high-level context and refine features dynamically, we use a gated encoding mechanism:

\begin{align}
Z &= W_1^{(1 \times 1)} * F_{\text{dual}} + b_1 \\
Z' &= \text{LeakyReLU}(Z) \\
\tilde{Z} &= W_2^{(1 \times 1)} * Z' + b_2 \\
G &= \sigma(\tilde{Z}) \\
F_{\text{encoded}} &= (F_{\text{dual}} + \tilde{Z}) \odot G
\end{align}

where $W^{(1 \times 1)}$ are 1x1 convolutions and $G$ is a learnable gating mechanism.

\subsection{Feature Fusion with Skip Connection}

To merge fine-grained low-level and abstract high-level features, we use a skip connection inspired by U-Net:

\begin{align}
F_{\text{resized}} &= \text{Interpolate}(F, \text{size} = (H_E, W_E)) \\
F_{\text{concat}} &= \text{Concat}(F_{\text{resized}}, F_{\text{encoded}}) \\
F_{\text{fused}} &= W_{\text{fusion}}^{(1 \times 1)} * F_{\text{concat}} + F_{\text{encoded}}
\end{align}

\subsection{Prediction}

We perform adaptive average pooling and flatten the feature map to obtain the final class logits:

\begin{align}
F_{\text{final}} &= \text{Flatten}(\text{AdaptiveAvgPool}(F_{\text{fused}})) \\
y &= W_{\text{pred}} \cdot F_{\text{final}} + b_{\text{pred}}
\end{align}

where $y \in \mathbb{R}^{B \times N}$ are the class logits.

\subsection{Summary of Model Computation}

The final pipeline is summarized as:

\begin{equation}
y = W_{\text{pred}} \cdot \text{Flatten} \left(
\text{AdaptiveAvgPool} \left(
W^{(1 \times 1)}_{\text{fusion}} \cdot \text{Concat}(F_{\text{encoded}}, \text{Interpolate}(F)) + F_{\text{encoded}}
\right) \right) + b_{\text{pred}}
\end{equation}

\section{Loss Function}

In this work, we propose a novel \textbf{Consistency Loss} that ensures stable feature learning and robust text recognition by integrating multiple loss components. Our loss function is designed to:

\begin{enumerate}
    \item Maintain consistency in attention across similar input samples.
    \item Regularize context encoding to prevent overfitting.
    \item Preserve feature integrity throughout the network.
\end{enumerate}

The total loss function is defined as:

\begin{equation}
\mathcal{L}_{total} = \lambda_{\text{att}} \mathcal{L}_{\text{att}} + \lambda_{\text{ctx}} \mathcal{L}_{\text{ctx}} + \lambda_{\text{fea}} \mathcal{L}_{\text{fea}} + \mathcal{L}_{\text{CE}}
\end{equation}

where:
\begin{itemize}
    \item $\mathcal{L}_{\text{att}}$ is the Attention Consistency Loss,
    \item $\mathcal{L}_{\text{ctx}}$ is the Context Regularization Loss,
    \item $\mathcal{L}_{\text{fea}}$ is the Feature Consistency Loss,
    \item $\mathcal{L}_{\text{CE}}$ is the standard Cross Entropy Loss,
    \item $\lambda_{\text{att}}, \lambda_{\text{ctx}}, \lambda_{\text{fea}}$ are hyperparameters controlling the weight of each component.
\end{itemize}

\subsection{Attention  Loss (TV Regularization)}

We define the attention consistency loss using Total Variation (TV) as follows:

\begin{equation}
\mathcal{L}_{\text{attn}} = \frac{1}{N} \sum_{i=1}^{N} TV(A_i)
\end{equation}

where $A_i$ is the attention map for the $i$-th sample, and the Total Variation is computed as:

\begin{equation}
TV(A) = \frac{1}{HW} \sum_{h=1}^{H-1} \sum_{w=1}^{W} |A_{h+1,w} - A_{h,w}| + \frac{1}{HW} \sum_{h=1}^{H} \sum_{w=1}^{W-1} |A_{h,w+1} - A_{h,w}|
\end{equation}

This loss penalizes large differences between neighboring pixels in both vertical and horizontal directions, encouraging smooth transitions in the attention maps. This is especially helpful in stroke-level recognition where attention should flow naturally along character contours.

\subsection{Context Regularization Loss}
Dynamic context encoding provides high-level contextual understanding of text features. However, excessive transformation may lead to loss of essential character details. To prevent this, we introduce a regularization term that constrainsencoded context from deviating too much from the original feature representation.

\begin{equation}
\mathcal{L}_{\text{ctx}} = \frac{1}{N} \sum_{i=1}^{N} \left\| C_i - C_i^{\text{orig}} \right\|_2^2
\end{equation}

where:
\begin{itemize}
    \item $C_i$ is the encoded context for sample $i$,
    \item $C_i^{\text{orig}}$ is the original feature map before context encoding.
\end{itemize}

\subsection{Feature Consistency Loss}

Feature consistency loss ensures that feature representations before and after edge-aware attention remain semanticallyconsistent. This prevents feature distortion caused by aggressive attention mechanisms.

\begin{equation}
\mathcal{L}_{\text{fea}} = \frac{1}{N} \sum_{i=1}^{N} \left\| F_i^{\text{dual}} - F_i^{\text{feat}} \right\|_2^2
\end{equation}

where:
\begin{itemize}
    \item $F_i^{\text{dual}}$ is the feature map after dual attention,
    \item $F_i^{\text{feat}}$ is the original backbone feature for sample $i$.
\end{itemize}

\subsection{Cross Entropy Loss}

In addition to the consistency losses, we apply standard Cross Entropy Loss for character classification:

\begin{equation}
\mathcal{L}_{\text{CE}} = - \frac{1}{N} \sum_{i=1}^{N} \sum_{c=1}^{C} \mathbb{1}_{[y_i = c]} \log p_{i,c}
\end{equation}

where:
\begin{itemize}
    \item $y_i$ is the ground-truth label,
    \item $p_{i,c}$ is the predicted probability for class $c$ of sample $i$,
    \item $\mathbb{1}_{[y_i = c]}$ is the indicator function.
\end{itemize}

\section{Dataset}

To evaluate the performance of the proposed Stroke-Sensitive Attention and Dynamic Context Encoding Network (SDA-Net), we introduce the \textbf{Among Car Plate Single Letter Dataset (ACPSLD)}. This large-scale dataset is specifically designed for single-character license plate recognition in real-world traffic environments. Unlike traditional OCR datasets, ACPSLD focuses on character-level extraction from vehicle license plates captured under dynamic and diverse conditions.

\subsection{Data Collection Method}

The dataset was collected from live CCTV footage recorded in real-world road environments, where moving vehicles are monitored under varying conditions such as lighting, weather, motion blur, and camera angle. The original dataset consists of:

\begin{itemize}
    \item 5,223 vehicle images for the training set,
    \item 974 vehicle images for the validation set,
    \item 391 vehicle images for the benchmarking set.
\end{itemize}

Each image contains one license plate. From these, individual characters were extracted and labeled to build a structured, single-character OCR dataset.

\subsection{Data Extraction and Annotation}

The character-level dataset was created through the following process:

\begin{enumerate}
    \item Automatic segmentation of license plate characters using a trained detection model.
    \item Manual verification and correction of labels for accuracy.
    \item Metadata annotation, including plate type, and character position within the plate.
\end{enumerate}

\subsection{Imbalance Handling Strategy}

A common issue in Korean license plate datasets is the imbalance between numeric and Korean alphabetic characters. Numeric digits appear significantly more often than letters, which may cause biased learning. To mitigate this, we employed the following strategies:

\begin{itemize}
    \item Equalized the number of samples between numeric and Korean characters.
    \item Applied targeted data augmentation (e.g., brightness and angle) on underrepresented classes.
    \item Ensured proportional inclusion of various license plate types in all splits.
\end{itemize}

This balancing strategy improves the model’s generalization across all character types and reduces performance discrepancies between numerals and letters.

\subsection{Dataset Structure}

The ACPSLD dataset is categorized by several attributes:

\begin{itemize}
    \item \textbf{Color Type:} White, green, blue, yellow, and black.
    \item \textbf{Usage Type:} Private, commercial, construction, and government vehicles.
    \item \textbf{Local:} State, City
    \item \textbf{Vehicle Type:} Car, Motorcycle
    \item \textbf{Format:} Standard, compressed, and specialized
    character plates.
\end{itemize}

Each character sample is labeled with:

\begin{itemize}
    \item Ground-truth text (a single character),
    \item Plate type metadata (e.g., usage type, state, city),
    \item Bounding box coordinates within the plate image.
\end{itemize}

\begin{table}[htbp]
\centering
\caption{Statistics of the ACPSLD dataset.}
\label{tab:acpsld}
\begin{tabular}{|l|c|}
\hline
\textbf{Dataset Split} & \textbf{Number of Vehicle Images} \\
\hline
Training Set & 5,223 \\
Validation Set & 974 \\
Benchmarking Set & 391 \\
\hline
\end{tabular}
\end{table}

\subsection{License Plate Types and Distribution}

\begin{figure}[htbp]
\centering
\includegraphics[width=0.9\textwidth]{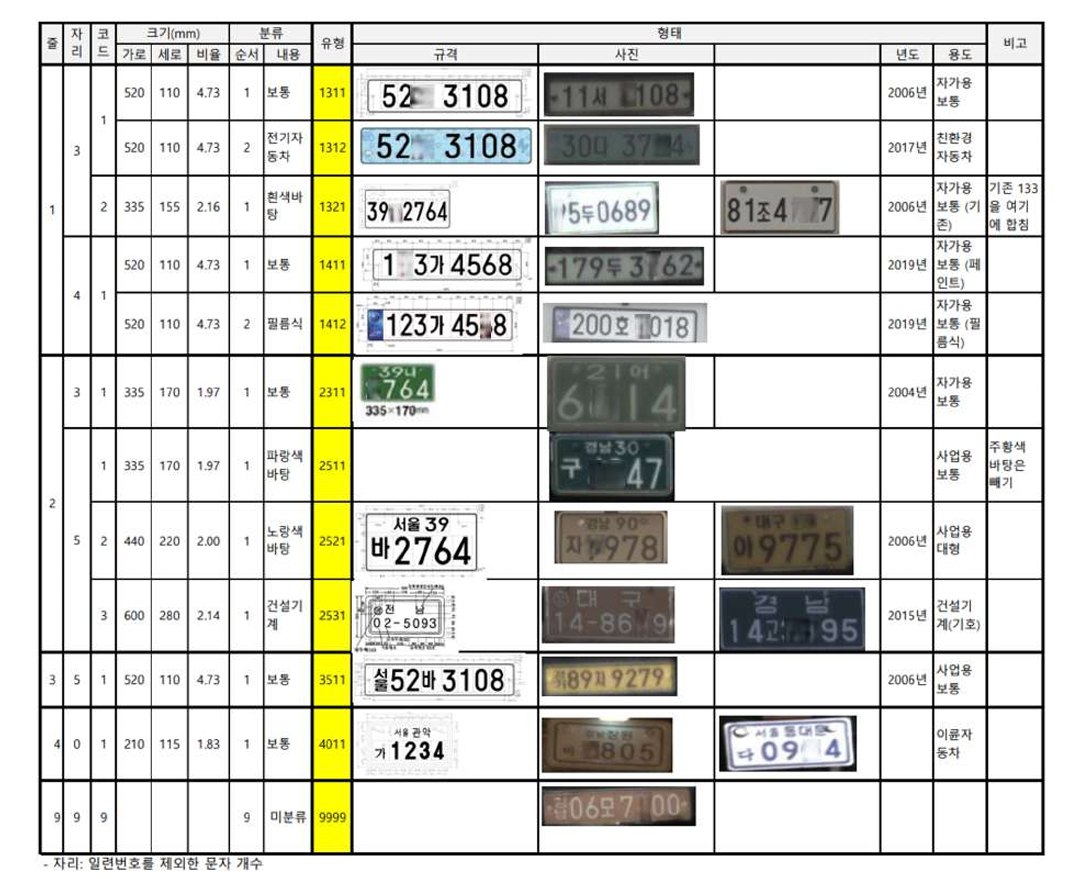}
\caption{Example license plate types and formats in the ACPSLD dataset.}
\label{fig:plate_types}
\end{figure}

Figure~\ref{fig:plate_types} illustrates various license plate types, colors, and formats included in ACPSLD, showing the diversity of the dataset.

\subsection{Implementation Details}

We implemented SDA-Net using PyTorch and trained the model on an NVIDIA GeForce RTX 3050 GPU with CUDA 11.8. The training hyperparameters are summarized in Table~\ref{tab:train_config}.

\begin{table}[htbp]
\centering
\caption{Training configuration for SDA-Net.}
\label{tab:train_config}
\begin{tabular}{|l|p{9cm}|}
\hline
\textbf{Hyperparameter} & \textbf{Value} \\
\hline
Optimizer & AdamW \\
Learning Rate & $5 \times 10^{-5}$ \\
Batch Size & 128 \\
Loss Function & Consistency Loss \\
Training Epochs & 100 \\
Data Augmentation & Random rotation ($\pm5^\circ$), brightness adjustment (0.9–1.1), Gaussian Blur, Contrast Adjustment \\
\hline
\end{tabular}
\end{table}

\section{Evaluation Methodology}

The evaluation of the proposed Stroke-Sensitive Attention and Dynamic Context Encoding Network (SDA-Net) is conducted in a real-world CCTV environment to verify its practical deployability. The methodology strictly follows the Korean National Police Agency (KNPA) standard for unmanned traffic enforcement equipment.

\subsection{Evaluation Criteria}

The evaluation adopts a correctness-based standard where:

\begin{itemize}
    \item A test case is considered \textbf{successful only if all characters} in a license plate are correctly recognized.
    \item Even a \textbf{single misclassification} leads to failure for the entire test case.
\end{itemize}

This strict metric reflects real-world deployment scenarios, where a single recognition error can result in incorrect citations or enforcement failures.

\subsection{Real-Time Deployment for Evaluation}

To ensure robustness, SDA-Net is deployed and tested on live CCTV feeds in actual traffic environments. The real-time evaluation pipeline is as follows:

\begin{enumerate}
    \item Vehicle images are captured from live CCTV streams.
    \item License plates are detected and cropped via object detection.
    \item Each character is segmented and passed to the OCR model.
    \item Recognized characters are concatenated to reconstruct the full plate.
    \item The reconstructed plate is compared against the ground-truth registration number.
\end{enumerate}

\subsection{Advantages of This Evaluation Method}

This evaluation strategy offers several benefits:

\begin{itemize}
    \item \textbf{Real-world validation:} Simulates actual usage scenarios in traffic enforcement.
    \item \textbf{Strict correctness requirement:} Emphasizes precision over per-character accuracy.
    \item \textbf{Regulatory alignment:} Fully compliant with KNPA specifications for automated enforcement systems.
\end{itemize}

\subsection{Evaluation Setup}

\begin{figure}[ht! ]
\centering
\includegraphics[width=0.9\textwidth]{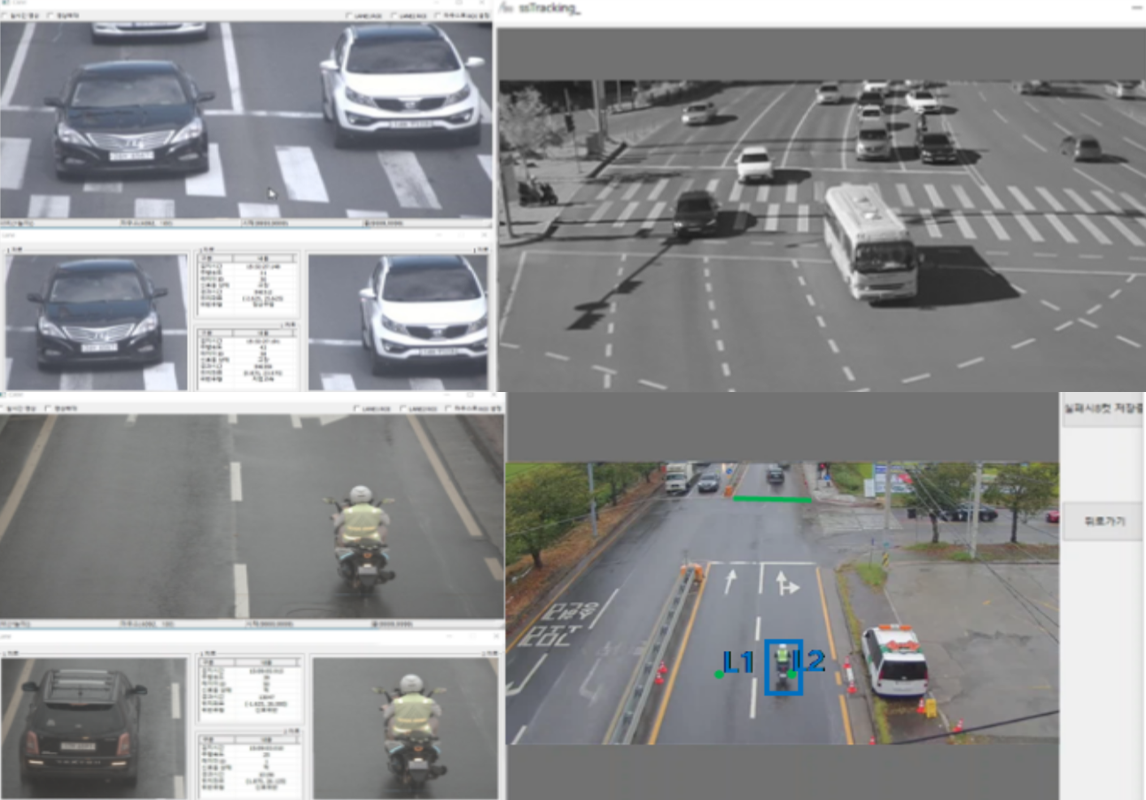}
\caption{Evaluation process for license plate recognition using real-time CCTV feed.}
\label{fig:evaluation_pipeline}
\end{figure}

Figure~\ref{fig:evaluation_pipeline} illustrates the real-time OCR evaluation process using surveillance CCTV.

\subsection{Real-Time On-Site Evaluation}

We conducted evaluation at various locations across Korea under different environmental conditions (day/night, urban/highway). Example license plates from actual footage are shown in Figure 3.

\begin{figure}[ht!]
\centering
\includegraphics[width=0.9\textwidth]{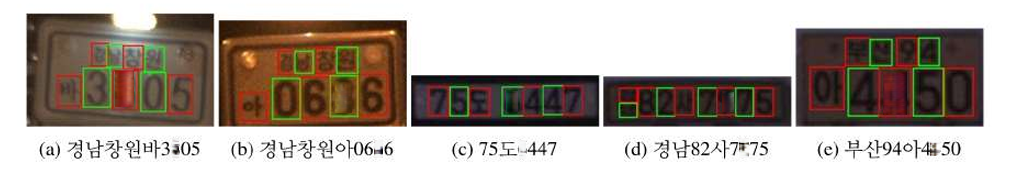}
\caption{Various license plates from on-site locations}
\label{fig:license_on_site}
\vspace{0.2em}
{\footnotesize \textit{*Due to personal data protection regulations of the Republic of Korea, parts of the license plate results cannot be publicly disclosed.}\par}
\end{figure}

Table~\ref{tab:realtime_results} summarizes real-time recognition results:

\begin{table}[ht! ]
\centering
\caption{Real-time on-site evaluation and ACPSLD benchmark results.}
\label{tab:realtime_results}
\begin{tabular}{|l|l|c|c|c|}
\hline
\textbf{Location} & \textbf{Environment} & \textbf{Total Vehicles} & \textbf{Recognized Vehicles} & \textbf{Recognition Rate (\%)} \\
\hline
Daegu Gamsam IC & Day/Night & 11,063 & 10,830 & 97.90 \\
Daegu Seongseo IC & Day/Night & 9,242 & 9,033 & 97.74 \\
Changwon Jangbuk-ro & Night & 431 & 388 & 90.02 \\
Changwon Metrocity & Day/Night & 101 & 98 & 97.03 \\
ACPSLD Benchmark & Day/Night & 391 & 354 & 90.54 \\
\hline
\end{tabular}
\end{table}

\subsection{Ablation Study}

To analyze the contribution of each module, we conducted an ablation study on the ACPSLD benchmark. Results are presented in Table~\ref{tab:ablation}.

\begin{table}[htbp]
\centering
\caption{Ablation study on ACPSLD benchmark dataset.}
\label{tab:ablation}
\begin{tabular}{|l|c|}
\hline
\textbf{Model Variant} & \textbf{Accuracy (\%)} \\
\hline
Baseline ResNet (No Attention) & 80.1 \\
+ Stroke-Sensitive Attention (SSA) & 84.7 \\
+ Edge-Aware Attention & 88.6 \\
+ Dynamic Context Encoding (DCE) & 90.5 \\
\hline
\end{tabular}
\end{table}

\section{Discussion}

While the real-time evaluation results of SDA-Net demonstrate consistently high accuracy exceeding 97\% across various deployment sites, performance on the ACPSLD Benchmark remains relatively lower at 90.54\%. This discrepancy can be explained by the design of the ACPSLD dataset, which deliberately includes a higher proportion of challenging and rare edge cases that are less frequently encountered in practical deployments.

\subsection{Factors Contributing to Benchmark Difficulty}

The lower recognition rate in the benchmark can be attributed to the following factors:

\begin{itemize}
    \item \textbf{Sequential Case Sampling:} The ACPSLD benchmark dataset is curated with samples arranged in increasing difficulty, introducing progressively complex challenges such as occlusion, poor lighting, and background clutter.
    
    \item \textbf{Difficult-to-Recognize Cases:} The dataset includes a high proportion of scenarios such as:
    \begin{itemize}
        \item Motorcycle license plates with smaller fonts and limited visibility.
        \item Plates covered in dust, dirt, or mud that obscure characters.
        \item Low-resolution images captured from long-distance surveillance.
        \item Partially occluded license plates due to structural elements or lighting reflections.
    \end{itemize}
    
    \item \textbf{Controlled Inclusion of Edge Cases:} Unlike real-world CCTV streams that predominantly contain clear, well-lit plates, the benchmark is designed to include rare but critical failure cases to test robustness.
\end{itemize}

\subsection{Significance of ACPSLD Benchmark}

Despite the drop in accuracy, the ACPSLD benchmark plays a vital role in enhancing OCR model robustness:

\begin{itemize}
    \item Improving performance on ACPSLD contributes to better generalization, enabling the model to recognize text accurately across a wide range of challenging environments.
    
    \item The benchmark encourages training on rare but practically important edge cases that might otherwise be underrepresented in real-time data.
    
    \item Optimization for this dataset ensures the model can operate reliably under fluctuating environmental factors in real-world deployments.
\end{itemize}

Therefore, although the recognition rate on ACPSLD is relatively lower, it serves as a highly effective benchmark for evaluating and enhancing the model's resilience and reliability in field applications.

\begin{figure}[htbp]
\centering
\includegraphics[width=0.9\textwidth]{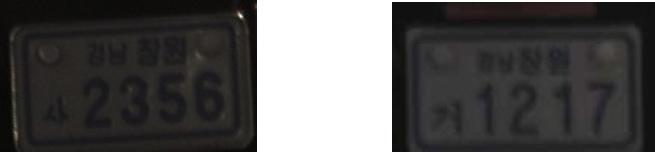}
\caption{Examples of difficult cases in the ACPSLD dataset, including dust-covered plates, occluded characters, and motorcycle license plates.}
\label{fig:acpsld_difficult_cases}
\end{figure}

\subsection{Overall Validation of SDA-Net}

The results validate the effectiveness of the proposed stroke-sensitive attention mechanism and dynamic context encoding module. SDA-Net demonstrates:

\begin{itemize}
    \item Strong generalization across both benchmark and live environments.
    \item Superior recognition performance compared to conventional lightweight OCR models.
    \item Practical applicability in traffic enforcement systems aligned with official standards.
\end{itemize}

These findings suggest that SDA-Net is well-suited for deployment in real-time intelligent surveillance systems requiring high accuracy and robustness.

\bibliographystyle{unsrt}
\bibliography{references}
\nocite{*}

\section*{Acknowledgments}
The authors would like to thank all members of Among Solution and Electro for their support.

\end{document}